%% file: llmConf_Arxiv.tex
\documentclass[a4paper]{article}
\usepackage{amsmath,amstext,amsgen,amsbsy,amsopn,amsfonts,amssymb}
\usepackage{easybmat}
\usepackage{graphics}
\usepackage{float} 
\usepackage[pdftex]{graphicx}
\usepackage{epsfig}
\usepackage{amsthm}
\usepackage{bm}
\usepackage{algorithm}
\usepackage{algorithmic}

\usepackage{wrapfig}
\usepackage{hyperref}
\usepackage{arydshln}
\usepackage{verbatim}
\usepackage{fancyvrb}
\usepackage{gensymb}
\begin{document}
\large
\title{\textbf{Mitigating Visual Hallucinations in Multimodal Systems through Retrieval-Augmented Reliability-Aware Inference}}
\author{
Pratheswaran Hariharan$^{\dag}$,  Haiping Xu$^{\dag }$, Donghui Yan$^{\ddag}$
\vspace{0.1in}\\
%%\vspace{0.1in}\\
$^{\dag}$Computer and Information Science\\[0.04in]
$^\ddag$Mathematics and Data Science\\[0.02in]
University of Massachusetts, Dartmouth, MA
%%$^\P$The Rivers School, Weston, MA%%\\[0.02in]
}

\date{\today}
\maketitle \normalsize
\begin{abstract} 
\noindent
Multimodal large language models (MLLMs) have demonstrated strong capabilities in vision-language understanding and natural-language response generation. However, these systems can still produce overconfident predictions and hallucination-like outputs, particularly when the visual evidence is weak, ambiguous, or semantically inconsistent. Most existing approaches focus on improving multimodal representation alignment or retrieval-augmented generation, while providing limited mechanisms to quantify instance-level prediction reliability or identify confidently incorrect visual outputs. To address this limitation, this work proposes a retrieval-augmented reliability-aware inference framework for trustworthy multimodal visual understanding. The proposed framework constructs an external visual evidence database using pretrained visual embeddings and nearest-neighbor retrieval over normalized feature representations. Retrieved evidence is used to estimate prediction trustworthiness through multiple reliability indicators, including similarity strength, class-support agreement, evidence margin, entropy-based uncertainty, and an aggregate reliability score. Based on these signals, a decision gate determines whether the system should accept the prediction, answer with caution, or abstain/fallback when evidence is insufficient. A multimodal response-generation layer then produces a final user-facing response conditioned on the reliability decision. Experiments on ImageNet-100 demonstrate that the proposed reliability-aware framework improves accepted prediction accuracy from 85.84\% to 88.88\% at 89.04\% coverage. The hallucination-like accepted wrong-answer rate is reduced from 14.16\% to 11.12\%. These results show that integrating retrieval evidence, reliability estimation, and selective decision gating can improve calibration and reduce overconfident visual errors without retraining large multimodal models.
\end{abstract}
%%\vspace{0.3cm}
%%\textbf{KEYWORDS:} Multimodal large language models, visual hallucination, reliability-aware inference, retrieval-augmented inference, evidence retrieval, confidence calibration, uncertainty estimation, selective prediction.
%%\end{tabular}

%%\begin{multicols}{2}

\input{introduction}

\input{related_work}

\input{methodology}

\input{experiments_results}

\input{limitation}

\input{discussion_conclusion}

\section*{Acknowledgment}
We thank the editors and referees for their careful review of this paper and the many valuable suggestions. This work is partially supported by the Office of Naval Research (ONR) Grant, MUST IV, UMass Dartmouth.

%%\bibliographystyle{plain}
%%\bibliographystyle{abbrv}
%%\bibliography{./jenrs}

\end{document}

%% file: introduction.tex
\section{Introduction}
Multimodal Large Language Models (MLLMs) have rapidly advanced vision-language understanding by enabling artificial intelligence systems to interpret visual inputs and generate natural-language responses. These models combine visual encoders with large language models to support tasks such as image description, visual question answering, cross-modal retrieval, and multimodal reasoning. Their ability to process heterogeneous information has made them increasingly useful in applications where users expect systems to explain, classify, or reason about visual content in an accessible language format. 
Despite these advances, current MLLMs still face reliability challenges. A major concern is that these systems may produce fluent and confident responses even when the visual evidence is weak, ambiguous, or inconsistent with the generated answer. Because the response is expressed in natural language, users may interpret it as reliable even when the underlying prediction is incorrect. This behavior is closely related to visual hallucination, where the system generates or accepts content that is not sufficiently grounded in the input image. In practical settings, such errors are more harmful when they are produced with high confidence, because users are less likely to question the output. 
\\
\\
Many existing systems focus primarily on improving representation alignment, language generation quality, or retrieval augmentation. While these directions improve the genera capability of MLLMs, they do not fully address the problem of instance-level reliability. In particular, a model may still be required to provide an answer for every input, regardless of whether the prediction is well supported. Internal confidence scores or softmax probabilities are often used as indicators of certainty, but these scores can be poorly calibrated and may not reflect actual correctness. As a result, a high-confidence prediction can still be wrong, especially for visually similar classes, uncertain samples, or inputs that do not strongly match the training distribution.
\\
\\
This creates an important gap in trustworthy multimodal inference: visual systems need mechanisms not only to predict an answer, but also to evaluate whether the answer should be trusted. A reliable system should be able to distinguish between cases where the evidence strongly supports a prediction and cases where the evidence is insufficient or conflicting. Instead of forcing a confident response for every image, the system should be capable of accepting reliable predictions, expressing caution when uncertainty exists, and abstaining or falling back when the available evidence does not justify a definitive answer. 
\\
\\
To address this problem, this paper proposes a retrieval-augmented reliability-aware inference for mitigating overconfident visual hallucinations in multimodal systems. The proposed approach introduces an external evidence-checking layer between visual prediction and final response generation. For each input image, pretained visual embeddings are extracted and compares against a large scale reference evidence database. The system retrieves visually similar examples and analyzes their consistency with the predicted class. This retrieved evidence provides an external grounding signal that helps determine whether the prediction is supported by similar visual instances. 
\\
\\
The reliability estimation module combines multiple evidence indicators rather than relying on a single confidence value. These indicators include similarity strength, class-support agreement, evidence margin, entropy-based uncertainty, and an aggregate reliability score. Similarity strength measures how closely the query image matches the retrieved examples. Class-support based agreement evaluates whether the retrieved neighbors consistently support the same predicted class. The margin of evidence captures the separation between the strongest and competing class supports, while entropy measures uncertainty in the distribution of the retrieved evidence. Together, these signals provide a more informative estimate of the trustworthiness of the prediction than raw confidence alone.
Based on the estimated reliability, the framework applies a selective decision gate with three possible outcomes: accept, answer with caution, or abstain/fallback. If the evidence is strong and consistent, the system accepts the prediction and allows a confident response. If the evidence is partially supportive but uncertain, the response is softened using a cautious language. If the retrieved evidence is weak or conflicting, the system avoids producing a definitive answer and instead abstains or triggers fallback behavior. This design reduces the risk of unsupported confident outputs while maintaining useful coverage on reliable inputs. 
The proposed framework is evaluated on ImageNet-100 using a retrieval-based visual evidence database and a validation set for quantitative assessment. The results show that reliability-aware gating improves accepted prediction accuracy, reduces hallucination-like accepted wrong predictions, lowers Expected Calibration Error, and decreases confident wrong cases compared with baseline retrieval system These findings suggest that retrieval evidence and selective decision control can serve as practical post-hoc reliability mechanism for multimodal visual inference without requiring retaining of large multimodal models.
\\
\\
The main contributions of this work are as follows. First, this paper proposes a retrieval-augmented reliability aware-inference framework that validates visual predictions using external visual evidence before generating final responses. Second, it introduces a multi-signal reliability estimation strategy that combines similarity strength, class-support consistency, margin, entropy, and aggregate reliability scoring. Third, it implements a selective decision mechanism that enables the system to accept reliable predictions, answer cautiously under moderate uncertainty, and abstain or fallback when evidence is insufficient. Finally, experimental evaluation demonstrates that the proposed framework improves accepted accuracy, reduces overconfident visual errors, and improves calibration compared to a baseline retrieval-based inference system.  

%% file: related_work.tex
\section{Related Work}
Multimodal Large Language Models (MLLMs) have become an important research direction because they combine visual perception with language-based reasoning. Earlier vision-language models focused on learning joint image-text representations for tasks such as image classification, image captioning, cross-modal retrieval, and visual question answering. Contrastive image-text pretraining methods such as CLIP demonstrated that visual concepts can be aligned with natural-language supervision and transferred effectively to downstream recognition tasks \cite{radford2021learning}. More recent multimodal systems, including LLaVA, BLIP-2, InstructBLIP, Flamingo, and MiniGPT-4, extend this capability by connecting pretrained visual encoders with large language models for open-ended visual instruction following and natural-language response generation \cite{liu2023visual,li2023blip,dai2023instructblip,alayrac2022flamingo,zhu2024minigpt}.
\\
\\
Although these models have shown strong performance across vision-language tasks, their fluency does not always guarantee reliability. Since the final output is often expressed in natural language, users may interpret the response as grounded and trustworthy even when the underlying visual evidence is insufficient. This creates a critical challenge for applications where the system is expected not only to generate a plausible response, but also to determine whether the response is visually supported. Therefore, reliability estimation has become an important requirement for trustworthy multimodal inference.
\\
\\
A major limitation of current MLLMs is visual hallucination, where the generated response includes objects, attributes, relationships, or semantic claims that are not sufficiently supported by the input image. Early studies on image captioning showed that models can hallucinate objects even when they are absent from the scene \cite{rohrbach2018object}. More recent studies have demonstrated that hallucination remains a persistent issue in large vision-language models, especially when the image contains ambiguous visual cues, fine-grained categories, or semantically similar objects \cite{li2023evaluating,bai2024hallucination,huang2024visual,leng2024mitigating}.
\\
\\
The problem becomes more serious when hallucinated or incorrect outputs are produced with high confidence. In practical settings, an uncertain answer may be acceptable if the system clearly communicates uncertainty. However, an incorrect answer presented as confident can mislead users and reduce trust in the model. Many multimodal systems are designed to answer every input, even when the available evidence is weak or conflicting. As a result, there is a need for inference-time mechanisms that can identify unsupported or confidently incorrect visual predictions before they are converted into final user-facing responses.
\\
\\
Retrieval-augmented methods provide a useful direction for improving grounding and interpretability. In natural language processing, retrieval-augmented generation uses external documents or memory to support generated answers \cite{lewis2020retrieval,guu2020retrieval}. Similar principles can be applied to visual inference by retrieving nearest-neighbor examples from a reference image database. In this setting, retrieved examples serve as external visual evidence that can be compared with the query image. Deep nearest-neighbor methods have shown that examples retrieved from representation space can provide interpretable support for model predictions \cite{papernot2018deep}. Efficient similarity search tools such as FAISS \cite{johnson2019billion} and the random projection forests based kNN search algorithm \cite{rpForests2021} make it possible to retrieve evidence from large-scale embedding databases during inference.
\\
\\
However, retrieval alone does not guarantee reliability. A system may retrieve visually similar examples, but the evidence may still be weak, mixed across multiple classes, or semantically ambiguous. Therefore, retrieved evidence must be evaluated before it is used to support a final prediction. This motivates the use of evidence-based reliability estimation to determine whether the retrieved examples provide reliable support for the predicted class.
\\
\\
Confidence calibration is closely related to trustworthy inference. Modern neural networks are often miscalibrated, meaning that their predicted confidence scores do not always match empirical correctness \cite{guo2017calibration}. A model may assign high confidence to an incorrect prediction, especially under visual ambiguity, class similarity, or distribution shift. Calibration methods such as temperature scaling aim to reduce this mismatch by aligning predicted probabilities with observed accuracy. Expected Calibration Error (ECE) is commonly used to measure the gap between confidence and correctness across confidence intervals.
\\
\\
Uncertainty estimation methods provide additional tools for identifying unreliable predictions. Bayesian approximations, dropout-based uncertainty, deep ensembles, and related approaches estimate when a model is likely to be uncertain or wrong \cite{gal2016dropout,lakshminarayanan2017simple,ovadia2019can}. While these methods are valuable, they may require architectural modifications, additional training, or multiple forward passes. For large multimodal systems, such requirements may be computationally expensive. In contrast, retrieval-based reliability estimation can be implemented as a post-hoc inference layer that evaluates prediction trustworthiness without retraining the underlying multimodal model.
\\
\\
Selective prediction allows a model to abstain when confidence or evidence is insufficient \cite{geifman2017selective}. Instead of forcing a prediction for every input, selective systems trade coverage for reliability. This principle is important for trustworthy artificial intelligence because abstaining from uncertain cases is often preferable to producing confident errors. Existing selective classification methods typically rely on confidence thresholds or uncertainty estimates. However, in multimodal visual inference, decision making can be strengthened by incorporating external evidence consistency. A reliability-aware decision gate can evaluate whether retrieved examples strongly support the predicted class, whether competing classes receive similar support, and whether the evidence distribution is uncertain.
\\
\\
The proposed framework builds on prior work in multimodal learning, visual hallucination analysis, retrieval-augmented inference, calibration, uncertainty estimation, and selective prediction. However, it focuses on a specific gap: inference-time reliability control for multimodal visual systems. Unlike methods that primarily improve representation learning or language generation, this work introduces an evidence-aware decision layer that validates visual predictions before producing the final response. Different from approaches that rely only on softmax confidence, the proposed framework estimates reliability using retrieved visual evidence, similarity strength, class-support agreement, evidence margin, entropy-based uncertainty, and an aggregate reliability score. Unlike standard retrieval-augmented methods, retrieved examples are not used only as supporting context; they are explicitly used to decide whether the system should accept, answer with caution, or abstain/fallback. This combination of retrieval evidence, reliability estimation, and selective response control provides a practical post-hoc mechanism for reducing overconfident visual errors in multimodal inference.

%% file: methodology.tex
\section{Methodology}
The proposed framework introduces a retrieval-augmented reliability-aware inference pipeline for mitigating overconfident visual hallucination-like errors in multimodal visual systems. The objective is not only to predict the most likely visual class, but also to determine whether the prediction is sufficiently supported by retrieved visual evidence before generating the final response. This design separates prediction generation from prediction acceptance. A class may be retrieved as the nearest semantic match, but it is treated as a reliable output only when the evidence gate confirms that the prediction is trustworthy.
\\
\\
To improve readability, the methodology is organized into three parts. First, the data source and preprocessing pipeline are described, including the ImageNet-100 dataset, image preprocessing, feature extraction, and embedding normalization. Second, the proposed reliability-aware inference framework is presented, including external evidence retrieval, class-support estimation, reliability signal computation, and decision-gated response control. Third, the evaluation protocol is described, including the calculation of accepted accuracy, coverage, abstention rate, calibration error, and hallucination-like accepted wrong-answer rate.

\subsection{Data Source and Preprocessing Pipeline}
\subsubsection{Dataset and Evidence Database Split}
The experiments use ImageNet-100 as a controlled visual recognition dataset. The ImageNet-100 training split is used to construct the external visual evidence database, while the validation split is used to evaluate the reliability-aware inference pipeline. In this work, the reference evidence database contains approximately 130,000 ImageNet-100 training images, and the quantitative validation is performed on 5,000 validation images.
\\
\\
The purpose of using ImageNet-100 is to evaluate the proposed framework in a controlled classification setting where the reference evidence database and validation labels are available. This setting allows the system to measure whether a prediction is correct, whether it is accepted by the reliability gate, and whether accepted wrong predictions are reduced after applying evidence-based reliability control.
\subsubsection{Image Preprocessing}
Each input image is converted to RGB format and processed using the same transformation pipeline used for the pretrained ResNet-50 model. The image is resized, converted into a tensor, and normalized using ImageNet statistics. The same preprocessing pipeline is applied to both reference images and query images so that all embeddings are produced under consistent conditions.
\\
\\
This consistency is important because the framework compares query embeddings against stored reference embeddings. If different preprocessing pipelines were used for the reference database and the query image, similarity scores could become unreliable. Therefore, the same image preprocessing operations are used throughout database construction, validation, and user-upload inference.
\subsubsection{Visual Feature Extraction and Normalization}
The framework uses a pretrained ResNet-50 model as the visual feature extractor. The final classification layer is removed, and the feature vector from the penultimate layer is used as the image representation. For an input image $x$, the encoder produces a 2048-dimensional embedding:
\begin{equation}
z = f_{\theta}(x), \quad z \in \mathbb{R}^{2048},
\end{equation}
where $f_{\theta}$ denotes the pretrained ResNet-50 feature extractor and $z$ is the extracted visual embedding.
\\
\\
The extracted embedding is then L2-normalized:
\begin{equation}
\tilde{z} = \frac{z}{\|z\|_2 + \epsilon},
\end{equation}
where $\epsilon$ is a small constant for numerical stability. Normalization ensures that similarity search is based on the direction of the embedding vector rather than its magnitude. Since all embeddings are normalized, nearest-neighbor retrieval is equivalent to cosine-similarity-based search.
\subsection{Proposed Reliability-Aware Inference Framework}
\subsubsection{Framework Overview}
\begin{figure}[t]
    \centering
    \includegraphics[width=0.90\textwidth]{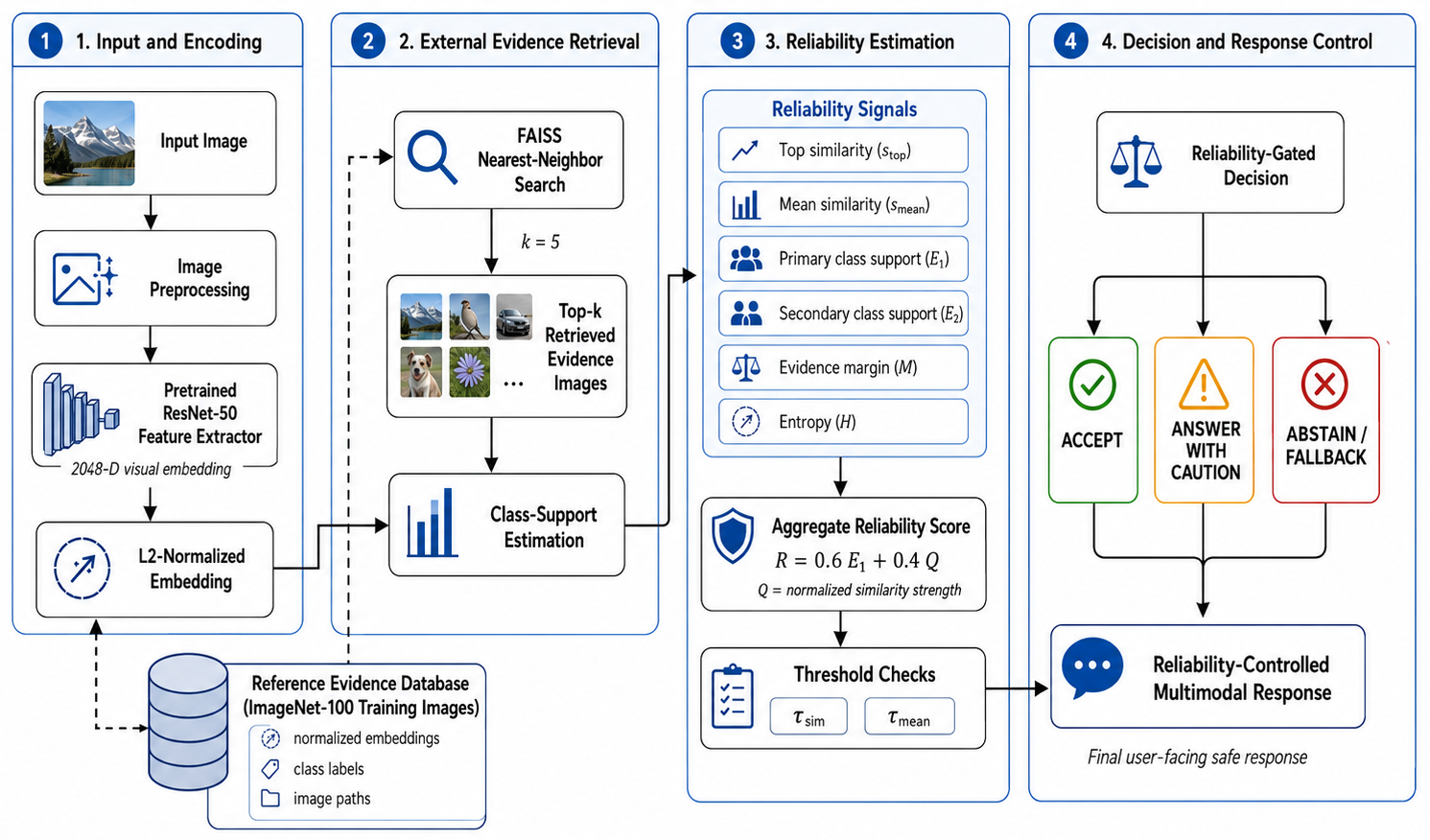}
    \caption{Overall architecture of the proposed retrieval-augmented reliability-aware inference framework.}
    \label{fig:overall architecture}
\end{figure}

Given an input image, the framework first converts the image into a dense visual embedding using the pretrained visual encoder. The embedding is then compared against a reference evidence database constructed from ImageNet-100 training images. The system retrieves the top-$k$ nearest visual examples and uses them as external evidence for the predicted class. Instead of directly trusting the nearest retrieved class, the framework evaluates the quality, strength, and consistency of the retrieved evidence.
\\
\\
As shown in Figure~\ref{fig:overall architecture}, the proposed framework follows a multi-stage reliability-aware inference pipeline. The input image is encoded into a normalized visual embedding, compared with an external evidence database using FAISS-based retrieval, evaluated using evidence-based reliability signals, and finally passed through a decision gate that controls whether the system should accept, answer with caution, or abstain/fallback.

%%
%%\\
%%\\
\noindent
\\
The overall framework consists of four major stages: input encoding, external evidence retrieval, reliability estimation, and decision-controlled response generation. The retrieved evidence is analyzed using multiple reliability indicators, including top similarity, mean similarity, primary class support, secondary class support, evidence margin, entropy-based uncertainty, and an aggregate reliability score. These indicators are passed to a decision gate that assigns the input image to one of three states: ACCEPT, ANSWER WITH CAUTION, or ABSTAIN/FALLBACK.
\\
%%\\
This architecture is designed to reduce unsupported confident predictions. If the retrieved examples strongly and consistently support the predicted class, the system provides a confident answer. If the evidence is partially supportive but uncertain, the response is softened using caution-aware language. If the evidence is weak or conflicting, the system avoids forcing a definitive prediction and instead abstains or falls back.
To summarize the complete inference process, Algorithm 1 presents the proposed retrieval-augmented reliability-aware inference procedure. The algorithm shows how the input image is encoded, compared with the external evidence database, evaluated using reliability signals, and finally passed through a decision gate before producing the final response.

\par\smallskip\smallskip
\noindent
\begin{minipage}{\columnwidth}
\footnotesize
\setlength{\fboxsep}{4pt}
\noindent\fbox{%
\begin{minipage}{0.94\columnwidth}
\textbf{Algorithm 1: Retrieval-Augmented Reliability-Aware Inference Framework}

\vspace{1mm}
\textbf{Input:} Query image $x_q$, reference evidence database $\mathcal{R}$, visual encoder $f_{\theta}$, number of neighbors $k$, thresholds $\tau_{\mathrm{sim}}$ and $\tau_{\mathrm{mean}}$.

\textbf{Output:} Final decision $D(x_q)$ and reliability-controlled response.

\vspace{1mm}
\begin{tabular}{@{}p{0.06\linewidth}p{0.88\linewidth}@{}}
1. & Preprocess the query image $x_q$ using the ImageNet-compatible transformation pipeline. \\[0.5mm]

2. & Extract the visual embedding using the pretrained encoder: $z_q=f_{\theta}(x_q)$. \\[0.5mm]

3. & Normalize the embedding: $\tilde{z}_q=z_q/(\|z_q\|_2+\epsilon)$. \\[0.5mm]

4. & Retrieve the top-$k$ nearest reference images from $\mathcal{R}$ using FAISS: $E_k(x_q)=\{(y_j,s_j,p_j)\}_{j=1}^{k}$. \\[0.5mm]

5. & Compute similarity-based evidence weights: $w_j=\exp(\alpha s_j)/\sum_{m=1}^{k}\exp(\alpha s_m)$. \\[0.5mm]

6. & Aggregate class-support scores: $S(c)=\sum_{j\in I_c}w_j$, and predict $\hat{y}=\arg\max_c S(c)$. \\[0.5mm]

7. & Compute reliability signals: $s_{\mathrm{top}}$, $s_{\mathrm{mean}}$, $E_1$, $E_2$, $M$, $H$, and $R$. \\[0.5mm]

8. & Evaluate the evidence gate: $\mathrm{GateFail}(x_q)=(s_{\mathrm{top}}<\tau_{\mathrm{sim}})\lor(s_{\mathrm{mean}}<\tau_{\mathrm{mean}})$. \\[0.5mm]

9. & If $\neg \mathrm{GateFail}(x_q)$, $R\geq0.85$, $H\leq0.30$, and $M\geq0.50$, assign $D(x_q)=\mathrm{ACCEPT}$. \\[0.5mm]

10. & Else if $R\geq0.70$ and $M\geq0.25$, assign $D(x_q)=\mathrm{ANSWER\ WITH\ CAUTION}$. \\[0.5mm]

11. & Otherwise, assign $D(x_q)=\mathrm{ABSTAIN/FALLBACK}$. \\[0.5mm]

12. & Generate the final response according to the selected decision state. \\
\end{tabular}
\end{minipage}%
}
\end{minipage}
%%\par\smallskip
%%\\
\\
The algorithm highlights that the proposed framework does not directly trust the nearest retrieved class. Instead, the final response is controlled by evidence strength, class-support consistency, uncertainty, and reliability score. This makes the inference process selective and reduces the chance of returning unsupported confident visual predictions.
\subsubsection{Operational Definition of Reliability}
In this study, reliability refers to an instance-level operational assessment of the strength and internal consistency of the retrieved visual evidence. It is constructed from retrieval similarity, neighborhood class support, evidence margin, and entropy, and is used to determine whether an individual prediction should be accepted, communicated cautiously, or rejected.
\\
\\
Reliability is not used as a statistical confidence interval, hypothesis-test significance level, or posterior probability of correctness. A statistical confidence interval quantifies uncertainty in an estimated population quantity, while a significance test evaluates evidence against a specified null hypothesis. In contrast, the retrieval-based reliability score is used during inference to control the response to a single query image.
\\
\\
These concepts serve different purposes. Statistical confidence intervals and significance tests describe uncertainty in aggregate experimental conclusions, whereas the proposed reliability score supports a per-instance decision. Therefore, a high reliability score indicates that the retrieved neighborhood is strong and internally consistent under the proposed evidence criteria, but it does not guarantee that the prediction is correct.
\\
\\
The reliability indicators are not independent because they are derived from the same top-k retrieved neighborhood. The method can identify retrieval results that are weak or internally conflicting, but it cannot independently verify whether a highly consistent retrieved neighborhood is semantically correct. A systematic retrieval failure may therefore cause several reliability indicators to appear strong simultaneously. The proposed method should consequently be interpreted as a retrieval-quality- aware selective procedure rather than an independent correctness verifier.
\subsubsection{Reference Evidence Database Construction}
The ImageNet-100 training set is used to construct the external visual evidence database. Each training image is passed through the same preprocessing and ResNet-50 feature extraction pipeline. For every reference image, the system stores the normalized embedding, the class label, and the image path. The reference evidence database is represented as:
\begin{equation}
\mathcal{R} = \{(\tilde{z}_i, y_i, p_i)\}_{i=1}^{N},
\end{equation}
where $\tilde{z}_i$ is the normalized embedding of the $i$-th reference image, $y_i$ is its class label, $p_i$ is its image path, and $N$ is the total number of reference images.
\\
\\
The reference database provides external visual evidence during inference. Rather than evaluating a query image in isolation, the system compares it with known reference examples. This allows the framework to estimate whether a predicted class is visually supported by similar images from the reference distribution.
\subsubsection{FAISS-Based Evidence Retrieval}
To perform efficient nearest-neighbor search over the reference database, the normalized embeddings are indexed using FAISS. For a query image embedding $\tilde{z}_q$, the FAISS index retrieves the top-$k$ most similar reference examples:
\begin{equation}
E_k(x_q) = \{(y_j, s_j, p_j)\}_{j=1}^{k},
\end{equation}
where $y_j$ is the class label of the $j$-th retrieved image, $s_j$ is its similarity score with the query image, and $p_j$ is the corresponding image path.
\\
\\
In the implemented system, $k=5$ is used for the user-facing evidence display and reliability computation. The retrieved images serve two purposes. First, they provide class-level support for prediction. Second, they provide interpretable evidence that can be displayed to users, allowing them to inspect the visual examples used to support or reject the prediction.
\subsubsection{Retrieval-Based Class-Support Estimation}
After the top-$k$ images are retrieved, the system computes class-level evidence support. Since the nearest neighbors may belong to different classes, the prediction is not based only on the single nearest image. Instead, the framework aggregates the retrieved evidence by class.
\\
\\
A similarity-weighted softmax function is applied to the retrieved similarity scores:
\begin{equation}
w_j =
\frac{\exp(\alpha s_j)}
{\sum_{m=1}^{k}\exp(\alpha s_m)},
\end{equation}
where $w_j$ is the normalized evidence weight of the $j$-th retrieved image and $\alpha$ controls the sharpness of the weighting distribution. In the implemented notebook, $\alpha=20$ is used to emphasize stronger retrieved matches.
\\
\\
For each class $c$ appearing in the retrieved set, the class-support score is computed as:
\begin{equation}
S(c) = \sum_{j \in I_c} w_j,
\end{equation}
where $I_c$ is the set of retrieved neighbors assigned to class $c$. The predicted class is selected as the class with the highest support:
\begin{equation}
\hat{y} = \underset{c}{\arg\max}\ S(c).
\end{equation}
This approach makes the prediction depend on the consistency of retrieved evidence. If the strongest retrieved images belong to the same class, the support for that class becomes high. If the retrieved images are distributed across competing classes, the prediction is treated as less reliable.
\subsubsection{Reliability Signal Computation}
The reliability module computes multiple evidence signals from the retrieved set. These signals capture both evidence strength and evidence consistency.
\\
\\
The first signal is the top similarity score:
\begin{equation}
s_{\mathrm{top}} = s_1.
\end{equation}
This represents the similarity of the strongest retrieved image. A high value indicates that at least one reference image closely matches the query image.
\\
\\
The second signal is the mean similarity score:
\begin{equation}
s_{\mathrm{mean}} =
\frac{1}{k}\sum_{j=1}^{k}s_j.
\end{equation}
This measures the overall strength of the retrieved evidence set and prevents the system from relying only on a single strong neighbor when the remaining retrieved images may be weak.
\\
\\
The primary evidence support $E_1$ is defined as the support of the predicted class:
\begin{equation}
E_1 = S(\hat{y}).
\end{equation}
The secondary evidence support $E_2$ is defined as the highest support among competing classes:
\begin{equation}
E_2 = \max_{c \neq \hat{y}} S(c).
\end{equation}
\\
\\
The evidence margin is computed similarly as in \cite{TACOMA2012}:
\begin{equation}
M = E_1 - E_2.
\end{equation}
A larger margin indicates that the predicted class is clearly separated from competing classes. A smaller margin indicates ambiguity because another class receives similar support.
\\
\\
The framework also computes a similarity gap between the strongest and weakest retrieved neighbors:
\begin{equation}
G = s_1 - s_k.
\end{equation}
This value provides an additional measure of retrieval separation within the top-$k$ evidence set.
\\
\\
Entropy is used to estimate uncertainty in the evidence distribution:
\begin{equation}
H = -\sum_c S(c)\log(S(c)+\epsilon).
\end{equation}
Low entropy indicates that the retrieved evidence is concentrated around one class. High entropy indicates that evidence is distributed across multiple classes, suggesting ambiguity or conflict among retrieved neighbors.
\subsubsection{Reliability Score and Decision Gate}
The aggregate reliability score combines evidence support and similarity strength into a single instance-level trust estimate. In the implemented framework, the reliability score is computed as:
\begin{equation}
R = 0.6E_1 + 0.4Q,
\end{equation}
where $E_1$ is the primary class-support score and $Q$ is the normalized similarity strength. The similarity strength is computed as:
\begin{equation}
Q = \min\left(\frac{s_{\mathrm{mean}}}{\tau_{\mathrm{mean}}}, 1.0\right),
\end{equation}
where $\tau_{\mathrm{mean}}$ is the mean-similarity threshold.
\\
\\
The reliability score gives greater weight to evidence agreement than to raw similarity strength because class-consistent retrieved evidence is more important than a single strong visual match. A prediction is considered more trustworthy when the retrieved examples are both visually similar and semantically consistent with the predicted class. The weights 0.6 and 0.4 are treated as validation-tuned hyperparameters that prioritize class-support consistency while still preserving the influence of retrieval similarity. Alternative weight ratios can be evaluated through sensitivity analysis when the dataset, encoder, or reference database changes.
\\
\\
The framework uses two similarity thresholds to determine when evidence is too weak to support a confident prediction: a top-similarity threshold $\tau_{\mathrm{sim}}$ and a mean-similarity threshold $\tau_{\mathrm{mean}}$. These thresholds are estimated from validation evidence statistics by computing low-quantile values of the top similarity and mean similarity distributions. Specifically, the implemented notebook uses the lower 10\% quantile of validation retrieval strength to identify weak-evidence cases. This produces the following thresholds in the implemented experiment:
\begin{equation}
\tau_{\mathrm{sim}} = 0.8416,
\end{equation}
\begin{equation}
\tau_{\mathrm{mean}} = 0.8285.
\end{equation}
These threshold values are validation-derived hyperparameters and may be recalibrated when the reference database, encoder, or dataset changes. If either the strongest retrieved match or the mean evidence strength is below threshold, the system treats the prediction as unreliable and triggers abstention or fallback behavior.
\\
\\
The decision gate converts the reliability signals into one of three output states: ACCEPT, ANSWER WITH CAUTION, or ABSTAIN/FALLBACK. The gate first checks whether the evidence strength fails the similarity thresholds:
\begin{equation}
\mathrm{GateFail}(x) =
(s_{\mathrm{top}} < \tau_{\mathrm{sim}})
\lor
(s_{\mathrm{mean}} < \tau_{\mathrm{mean}}).
\end{equation}
Here, $\mathrm{GateFail}(x)$ indicates that the retrieved visual evidence for input image $x$ is insufficient to support a confident prediction. The gate fails when either the strongest retrieved match is below the top-similarity threshold or the average similarity of the retrieved evidence set is below the mean-similarity threshold.
\\
\\
If the evidence fails this condition, the system does not directly accept the prediction. The final decision is then based on the reliability score, entropy, and margin. In the implemented user-facing pipeline, the decision is assigned as follows:
\begin{equation}
D(x)=
\begin{cases}
A, & \neg \mathrm{GateFail}(x),\ R \geq 0.85,\ H \leq 0.30,\ M \geq 0.50,\\
C, & R \geq 0.70,\ M \geq 0.25,\\
F, & \mathrm{otherwise}.
\end{cases}
\end{equation}
where $A$ denotes ACCEPT, $C$ denotes ANSWER WITH CAUTION, and $F$ denotes ABSTAIN/FALLBACK. The accept state is used when the retrieved evidence is strong, class-consistent, and low in uncertainty. The caution state is used when evidence is partially supportive but not strong enough for full confidence. The abstain/fallback state is used when evidence is weak, conflicting, or insufficient.
\\
\\
This decision mechanism is central to the proposed framework. It prevents the system from treating all predictions equally and allows it to adapt its response based on evidence quality. As a result, the system can reduce confidently incorrect outputs while maintaining useful coverage on reliable cases.
\subsubsection{Reliability-Controlled Multimodal Response Generation}
After the reliability gate determines the decision state, the system generates the final user-facing response using a multimodal language model. In the implemented pipeline, LLaVA is used to generate a short visual description of the input image. To reduce confirmation bias, the model first describes the image without being directly conditioned on the retrieved class label.
\\
\\
The final response is then controlled using the reliability decision. If the decision is ACCEPT, the system reports the predicted class and indicates that the retrieved evidence is strong and consistent. If the decision is ANSWER WITH CAUTION, the system presents the predicted class as a possible interpretation and uses caution-aware language. If the decision is ABSTAIN/FALLBACK, the system avoids making a definitive class claim and explains that the retrieved evidence is weak, ambiguous, or conflicting.
\\
\\
This response-control step ensures that the language output follows the evidence-based reliability assessment rather than relying only on the nearest retrieved class or the generative model output. In the user-upload demo, the interface reports the predicted class, reliability score, top similarity, mean similarity, margin, entropy, decision label, retrieved evidence images, and final safe response. Since no ground-truth label is available for arbitrary uploaded images, the upload demo reports instance-level reliability, while overall performance is measured separately using the ImageNet-100 validation set.
\subsection{Evaluation Protocol}
The validation pipeline applies the retrieval and reliability mechanism to the ImageNet-100 validation set. For each validation image, the system retrieves the top-$k$ evidence images, predicts a class using class-support aggregation, computes reliability signals, and determines whether the sample is accepted or abstained.
\\
\\
The resulting validation records include the true label, predicted label, correctness, abstention status, reliability score, evidence support values, margin, entropy, top similarity, mean similarity, and similarity gap. These records are stored in a structured results table and used to compute overall performance.
\\
\\
The evaluation uses baseline accuracy, accepted accuracy, coverage, abstention rate, hallucination-like accepted wrong-answer rate, Expected Calibration Error (ECE), and confident wrong cases. The baseline retrieval system is evaluated as an always-answering system, meaning that it returns one predicted class for every validation image. Baseline accuracy is computed as:
\begin{equation}
Acc_{\mathrm{base}} = \frac{N_{\mathrm{correct}}}{N},
\end{equation}
where $N_{\mathrm{correct}}$ is the number of correct predictions and $N$ is the total number of validation samples.
\\
\\
For the proposed reliability-aware framework, accuracy is computed only over the samples accepted by the evidence gate. Therefore, accepted accuracy is defined as:
\begin{equation}
Acc_{\mathrm{accepted}} =
\frac{N_{\mathrm{accepted\ correct}}}{N_{\mathrm{accepted}}},
\end{equation}
where $N_{\mathrm{accepted\ correct}}$ is the number of accepted predictions that are correct and $N_{\mathrm{accepted}}$ is the total number of accepted predictions. Coverage is computed as:
\begin{equation}
Coverage = \frac{N_{\mathrm{accepted}}}{N},
\end{equation}
and abstention rate is computed as:
\begin{equation}
Abstention = \frac{N_{\mathrm{abstained}}}{N}.
\end{equation}
In this work, a hallucination-like visual error is defined as an incorrect visual prediction that is accepted by the system and therefore would be returned as a user-facing answer. This definition does not claim to measure all forms of generative hallucination. Instead, it focuses on the specific reliability problem studied in this paper: confidently accepted wrong visual predictions. For the always-answering baseline, the hallucination-like error rate is equivalent to the prediction error rate:
\begin{equation}
HL_{\mathrm{base}} =
\frac{N_{\mathrm{wrong}}}{N}
=
1 - Acc_{\mathrm{base}}.
\end{equation}
For the proposed framework, the hallucination-like accepted wrong-answer rate is computed only over accepted outputs:
\begin{equation}
HL_{\mathrm{accepted}} =
\frac{N_{\mathrm{accepted\ wrong}}}{N_{\mathrm{accepted}}}
=
1 - Acc_{\mathrm{accepted}}.
\end{equation}
This metric is reported separately from coverage because a selective system may improve accepted accuracy by abstaining on uncertain cases. Therefore, hallucination-like error rate should be interpreted together with coverage and abstention rate. A lower hallucination-like accepted wrong-answer rate indicates that fewer incorrect visual predictions are allowed to pass through the reliability gate as final accepted outputs.
\\
\\
Expected Calibration Error is computed using reliability scores as confidence estimates over 10 bins:
\begin{equation}
ECE =
\sum_{b=1}^{B}
\frac{|B_b|}{n}
\left|acc(B_b)-conf(B_b)\right|,
\end{equation}
where $B_b$ is the set of samples in bin $b$, $acc(B_b)$ is the empirical accuracy of the bin, $conf(B_b)$ is the average reliability score in the bin, and $n$ is the total number of samples. Confident wrong cases are defined as incorrect predictions with reliability greater than or equal to 0.85. The baseline retrieval system is treated as answering every sample, while the proposed reliability-aware framework only returns accepted predictions. This allows the evaluation to measure whether the proposed reliability gate reduces accepted wrong predictions and improves trustworthiness.

%% file: experiments_results.tex
\section{Experiments and Results}
%%
%%\begingroup
%%\captionsetup{font=footnotesize,skip=0pt}
%%\setlength{\abovecaptionskip}{1pt}
%%\setlength{\belowcaptionskip}{1pt}
%%
This section evaluates the proposed retrieval-augmented reliability-aware inference framework on the ImageNet-100 validation set and selected user-uploaded qualitative examples. The objective is to determine whether external visual evidence and reliability-aware decision gating can reduce overconfident wrong predictions while maintaining useful prediction coverage. The baseline retrieval system always returns a prediction for every validation image. In contrast, the proposed framework accepts a prediction only when the retrieved evidence is sufficiently strong, consistent, and reliable.
\\
\\
The evaluation focuses on three questions. First, does the proposed framework improve the accuracy of predictions that are accepted by the system? Second, does the reliability gate reduce hallucination-like accepted wrong answers and confident wrong cases? Third, do qualitative examples show that the framework behaves appropriately under reliable, ambiguous, weak-evidence, and out-of-coverage conditions?
\subsection{Experimental Setup}
The experiments were conducted using the ImageNet-100 dataset. The reference evidence database was constructed from approximately 130,000 ImageNet-100 training images, and quantitative validation was performed on 5,000 validation images. Each image was passed through a pretrained ResNet-50 feature extractor with the final classification layer removed, producing a 2048-dimensional visual embedding. All embeddings were $L_2$-normalized before indexing and retrieval.
\\
\\
FAISS was used to perform efficient nearest-neighbor search over the reference evidence database. For each validation image, the top-$k$ nearest reference examples were retrieved with $k=5$. The retrieved similarity scores were converted into class-support weights using a softmax weighting function with $\alpha=20$. The predicted class was selected based on the highest aggregated class support among the retrieved evidence examples.
\\
\\
The reliability gate used two validation-derived similarity thresholds: $\tau_{\mathrm{sim}}=0.8416$ for top similarity and $\tau_{\mathrm{mean}}=0.8285$ for mean similarity. A prediction was accepted only when the retrieved evidence was sufficiently strong and class-consistent. Otherwise, the system either abstained or triggered fallback behavior. The baseline retrieval system was evaluated as an always-answering system, while the proposed framework was evaluated based on accepted predictions.
\subsection{Evaluation Metrics}
The evaluation uses baseline accuracy, accepted accuracy, coverage, abstention rate, hallucination-like accepted wrong-answer rate, Expected Calibration Error (ECE), and confident wrong cases. Baseline accuracy measures the correctness of the retrieval prediction when the system answers every validation image. Accepted accuracy measures correctness only among predictions accepted by the reliability gate. Therefore, baseline accuracy and accepted accuracy are calculated on different output sets.
\\
\\
For the baseline retrieval system, accuracy is computed as the number of correct predictions divided by the total number of validation samples:
\begin{equation}
Acc_{\mathrm{base}} = \frac{N_{\mathrm{correct}}}{N}.
\end{equation}
Here, $N_{\mathrm{correct}}$ is the number of correct predictions and $N$ is the total number of validation samples.
\\
\\
For the proposed reliability-aware framework, accepted accuracy is computed only over accepted predictions:
\begin{equation}
Acc_{\mathrm{accepted}} =
\frac{N_{\mathrm{acc\_correct}}}{N_{\mathrm{acc}}}.
\end{equation}
Here, $N_{\mathrm{acc\_correct}}$ is the number of accepted predictions that are correct and $N_{\mathrm{acc}}$ is the total number of accepted predictions. Coverage is the proportion of validation samples accepted by the gate:
\begin{equation}
Coverage = \frac{N_{\mathrm{acc}}}{N},
\end{equation}
and abstention rate is the proportion of validation samples rejected by the gate:
\begin{equation}
Abstention = \frac{N_{\mathrm{abstained}}}{N}.
\end{equation}
\\
In this paper, a hallucination-like visual error is defined as an incorrect visual prediction that is accepted by the system and would therefore be returned as a user-facing answer. This definition does not measure all possible forms of generative hallucination. Instead, it focuses on the reliability problem studied in this work: accepted wrong visual predictions that may appear trustworthy to the user. For the always-answering baseline, every prediction is accepted, so the hallucination-like error rate is equivalent to the ordinary prediction error rate:
\begin{equation}
HL_{\mathrm{base}} =
\frac{N_{\mathrm{wrong}}}{N}
=
1-Acc_{\mathrm{base}}.
\end{equation}
For the proposed framework, the hallucination-like accepted wrong-answer rate is calculated only over predictions accepted by the reliability gate:
\begin{equation}
HL_{\mathrm{accepted}} =
\frac{N_{\mathrm{acc\_wrong}}}{N_{\mathrm{acc}}}
=
1-Acc_{\mathrm{accepted}}.
\end{equation}
\\
Thus, the hallucination-like error rate is related to prediction error, but it is interpreted differently for the selective framework because the system does not answer every sample. For this reason, hallucination-like error must be reported together with coverage and abstention rate. ECE is computed using the reliability score as the confidence estimate over 10 confidence bins. Confident wrong cases are defined as incorrect predictions with reliability greater than or equal to 0.85.
\subsection{Overall Quantitative Results}
Table~\ref{tab:overall_results} compares the baseline retrieval system with the proposed reliability-aware framework. In the table, the row labeled ``Proposed Reliability-Aware Framework'' refers to the complete framework proposed in this paper, including evidence retrieval, reliability estimation, and decision-gated response control. The baseline retrieval system uses the same retrieval-based prediction mechanism but does not apply selective reliability gating; therefore, it returns a prediction for every validation image.
\\
\\
The baseline retrieval system achieved an overall accuracy of 85.84\%, meaning that 14.16\% of the always-returned predictions were incorrect. Since the baseline accepts every prediction, its hallucination-like error rate is the same as its prediction error rate. After applying the reliability-aware evidence gate, the accepted prediction accuracy improved to 88.88\%. This shows that the proposed gate filters out a portion of unreliable predictions before they are returned as final answers.
\\
\\
The proposed framework accepted 89.04\% of the validation samples and abstained on 10.96\% of uncertain or weak-evidence cases. Although this reduces coverage, it improves the trustworthiness of the predictions actually returned to the user. The hallucination-like accepted wrong-answer rate decreased from 14.16\% to 11.12\%, corresponding to an absolute reduction of 3.04 percentage points and a relative reduction of 21.48\%. This result supports the main objective of the framework: reducing confidently accepted wrong visual outputs without retraining the underlying visual encoder or multimodal model.
%%
%%\par\noindent
%%\begin{minipage}{\columnwidth}
\begin{table}
\caption{\it Overall performance comparison between the baseline retrieval system and the proposed reliability-aware framework.}
\label{tab:overall_results}
\centering
\resizebox{\columnwidth}{!}{%
\begin{tabular}{lccccccc}
\hline
\textbf{System} & \textbf{Acc.} & \textbf{HL Error} & \textbf{Cov.} & \textbf{Abst.} & \textbf{ECE} & \textbf{Conf. Wrong} \\
 & \textbf{(\%)} & \textbf{(\%)} & \textbf{(\%)} & \textbf{(\%)} & \textbf{(\%)} &\textbf{(\%)}  \\
\hline
Baseline Retrieval & 85.84 & 14.16 & 100.00 & 0.00 & 7.40 & 263 \\
Proposed Reliability-Aware Framework & 88.88 & 11.12 & 89.04 & 10.96 & 5.87 & 226 \\
\hline
\end{tabular}%
}
\end{table}
%%\end{minipage}
Here, HL Error denotes the hallucination-like accepted wrong-answer rate. For the baseline retrieval system, this value is equal to the ordinary prediction error rate because the baseline always returns a prediction. For the proposed framework, this value is computed only over predictions accepted by the reliability gate. Therefore, HL Error should be interpreted together with coverage and abstention rate rather than as a standalone accuracy measure.
\subsection{Calibration and Reliability Analysis}
The reliability-aware gate also improved calibration. The baseline retrieval system produced an ECE of 7.40\%, while the proposed reliability-aware framework reduced ECE to 5.87\%. This reduction indicates that the reliability score better reflects empirical correctness after unreliable samples are filtered out. In practical terms, the system becomes less likely to present unsupported predictions as reliable outputs.
\\
\\
The number of confident wrong cases also decreased after applying the reliability gate. The baseline system produced 263 confident wrong predictions, while the proposed framework reduced this number to 226. This reduction is important because high-confidence wrong predictions are among the most harmful visual errors in user-facing multimodal systems. By rejecting weak or conflicting evidence cases, the proposed framework reduces the risk of misleading users with confident but incorrect outputs.
\subsection{Evidence Quality Analysis}
Table~\ref{tab:evidence_quality} summarizes the difference between accepted samples and abstained/fallback samples. Accepted samples achieved an accuracy of 88.88\%, while abstained/fallback samples had a much lower accuracy of 61.13\%. This confirms that the reliability gate is not rejecting samples randomly; instead, it tends to reject cases that are genuinely harder or less reliable.
\\
\\
Accepted samples also showed stronger evidence signals. Their average reliability score was 0.9560, compared with 0.8137 for abstained/fallback samples. The accepted group had higher average top similarity, higher mean similarity, larger evidence margin, and lower entropy. Specifically, accepted samples had an average margin of 0.8666 and entropy of 0.1563, while abstained/fallback samples had a lower margin of 0.5380 and a higher entropy of 0.6308. These results show that accepted predictions are supported by stronger and more class-consistent retrieved evidence, while abstained/fallback samples contain weaker or more ambiguous evidence.

%%\par\noindent
%%\begin{minipage}{\columnwidth}
\begin{table}
\caption{\it Evidence quality comparison between accepted and abstained/fallback samples.}
\label{tab:evidence_quality}
\centering
\resizebox{\columnwidth}{!}{%
\begin{tabular}{lccccccc}
\hline
\textbf{Group} & \textbf{Samples} & \textbf{Acc.} & \textbf{Rel.} & \textbf{$s_{\mathrm{top}}$} & \textbf{$s_{\mathrm{mean}}$} & \textbf{Margin} & \textbf{Entropy} \\
\hline
Accepted & 4452 & 0.8888 & 0.9560 & 0.9110 & 0.9000 & 0.8666 & 0.1563 \\
Abstained/Fallback & 548 & 0.6113 & 0.8137 & 0.8168 & 0.8028 & 0.5380 & 0.6308 \\
\hline
\end{tabular}%
}
\end{table}
%%\end{minipage}
%%
\subsection{Qualitative Case Analysis}
In addition to the quantitative evaluation, qualitative case analysis was conducted to examine how the proposed framework behaves at the individual-image level. This analysis is important because the objective of the proposed method is not only to improve aggregate accuracy, but also to control the final response based on the strength and consistency of retrieved evidence. A reliable multimodal visual system should accept predictions when evidence is strong, but should avoid confident answers when the retrieved evidence is weak, ambiguous, conflicting, or outside the reference class distribution.
\\
\\
Table~\ref{tab:qualitative_cases} summarizes four representative qualitative examples. These examples include one high-confidence accepted example and three abstain/fallback examples. The accepted case demonstrates that the proposed framework does not unnecessarily reject reliable predictions. The fallback cases demonstrate how the system handles ambiguous visual evidence, degraded image quality, and out-of-coverage inputs.

\begin{figure}
\centering
\includegraphics[width=0.63\columnwidth,trim=0.03in 0.03in 0.03in 0.03in,clip]{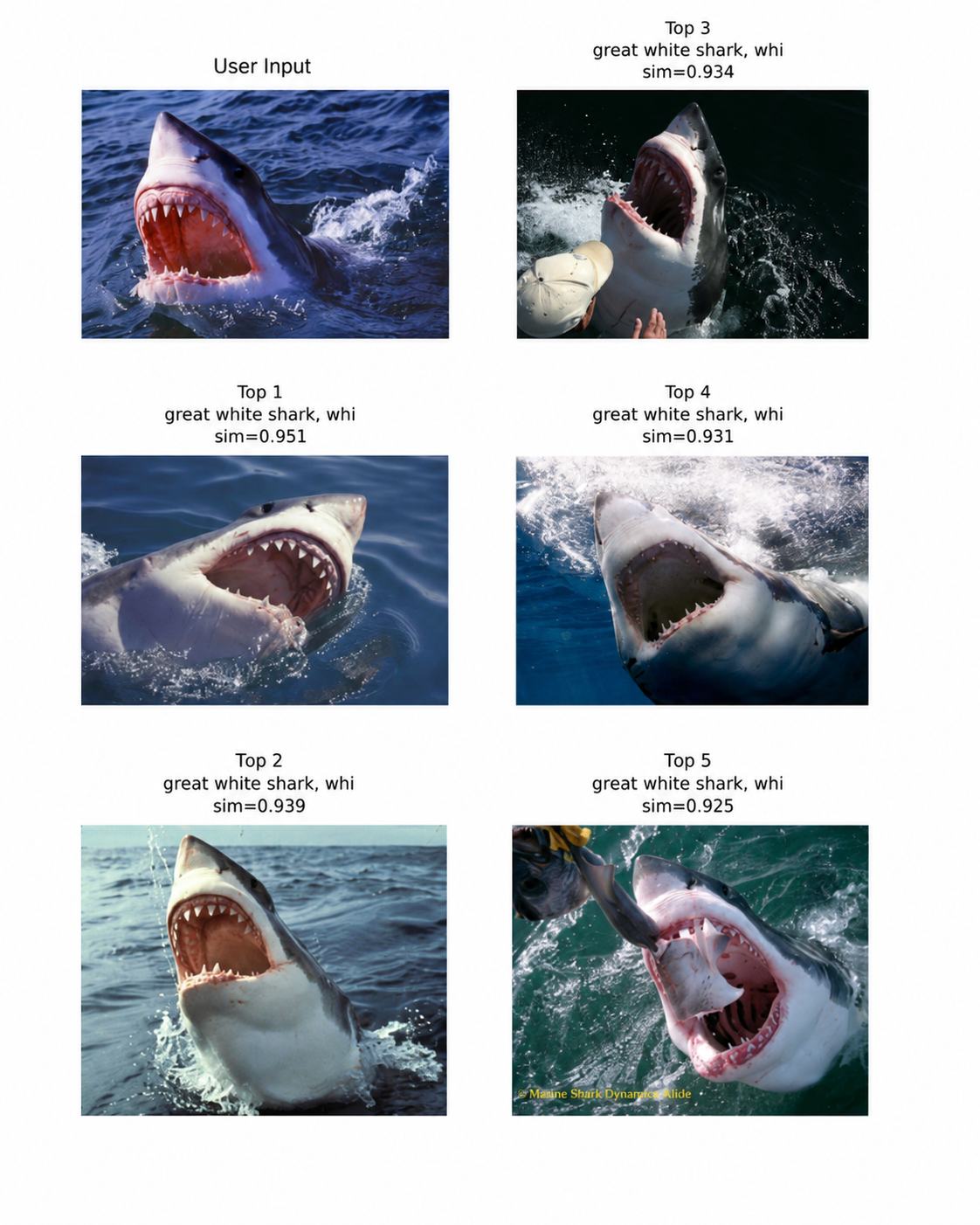}
\caption{\it Accepted qualitative case showing the input image and top-5 retrieved evidence for the great white shark example.}
\label{fig:shark_evidence}
\end{figure}

\begin{figure}
\centering
\includegraphics[width=0.80\columnwidth,trim=0.03in 0.03in 0.03in 0.03in,clip]{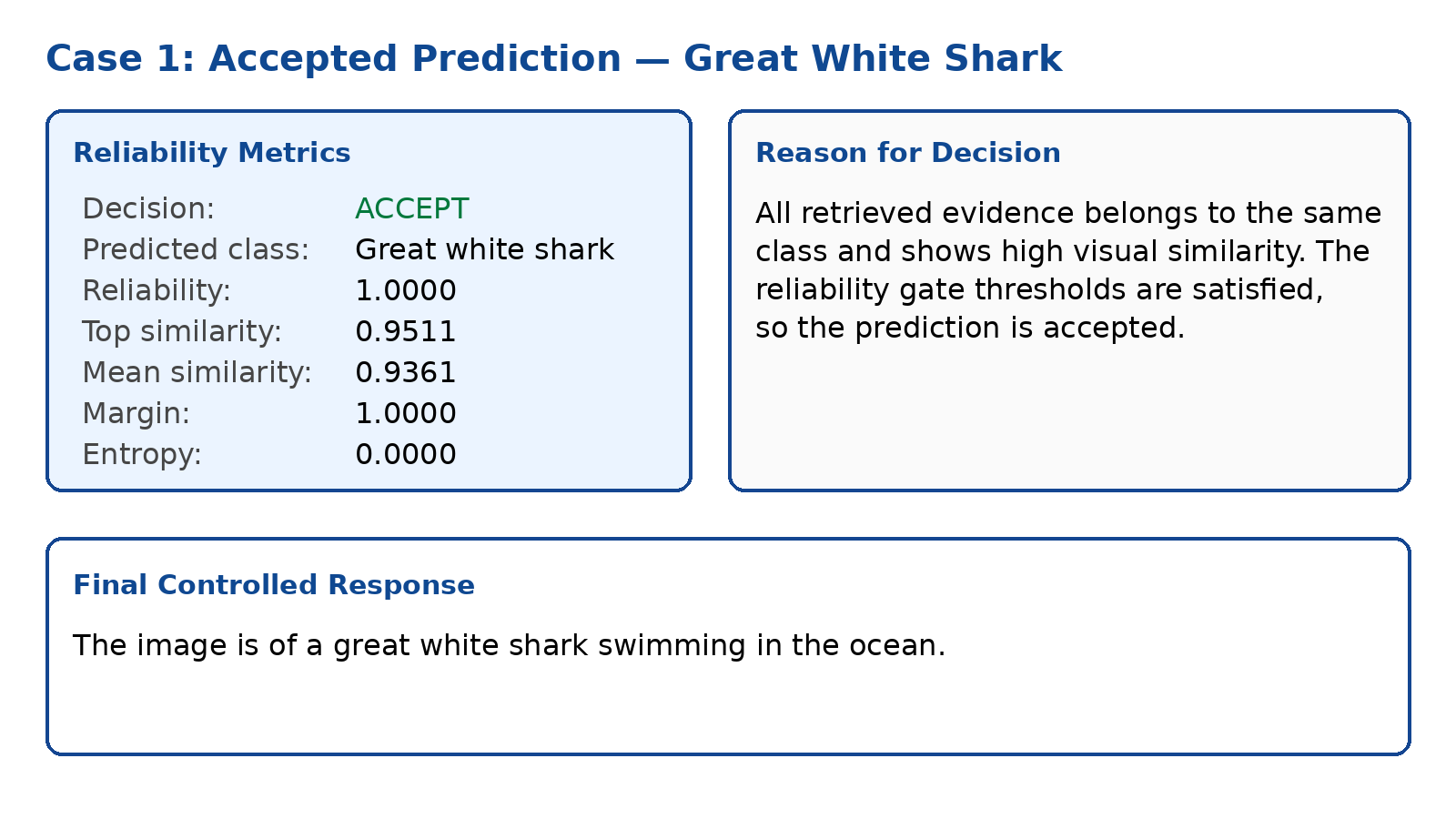}
\caption{\it Reliability summary and final controlled response for the accepted great white shark example.}
\label{fig:shark_summary}
\end{figure}

%%\par\noindent
%%\begin{minipage}{\columnwidth}
\begin{table}
\caption{\it Qualitative case study examples showing reliability-aware decision outcomes.}
\label{tab:qualitative_cases}
\centering
\resizebox{\columnwidth}{!}{%
\begin{tabular}{llllcl}
\hline
\textbf{Case} & \textbf{Input} & \textbf{Pred.} & \textbf{Decision} & \textbf{Rel.} & \textbf{Reason} \\
\hline
1 & Shark & Shark & ACCEPT & 1.0000 & Strong same-class evidence \\
2 & Birds & Kite & FALLBACK & 0.5931 & Mixed bird-class evidence \\
3 & Blurred car & Turtle & FALLBACK & 0.4316 & Weak unrelated retrieval \\
4 & Airplane & Crane & FALLBACK & 0.5505 & Outside reference coverage \\
\hline
\end{tabular}%
}
\end{table}
%%\end{minipage}
\noindent
\\
\textit{Case 1: Accepted prediction--great white shark.}
The first qualitative case, shown in Figure~\ref{fig:shark_evidence} and Figure~\ref{fig:shark_summary}, demonstrates a high-confidence accepted prediction. The input image contains a great white shark, and all top-five retrieved evidence examples belong to the same class. The reliability score is 1.0000, the top similarity is 0.9511, the mean similarity is 0.9361, the evidence margin is 1.0000, and the entropy is 0.0000. Since the retrieved examples are visually close, class-consistent, and free from competing class evidence, the decision gate accepts the prediction.

\par\noindent

\par\noindent
\begin{figure}
\centering
\includegraphics[width=0.63\columnwidth,trim=0.03in 0.03in 0.03in 0.03in,clip]{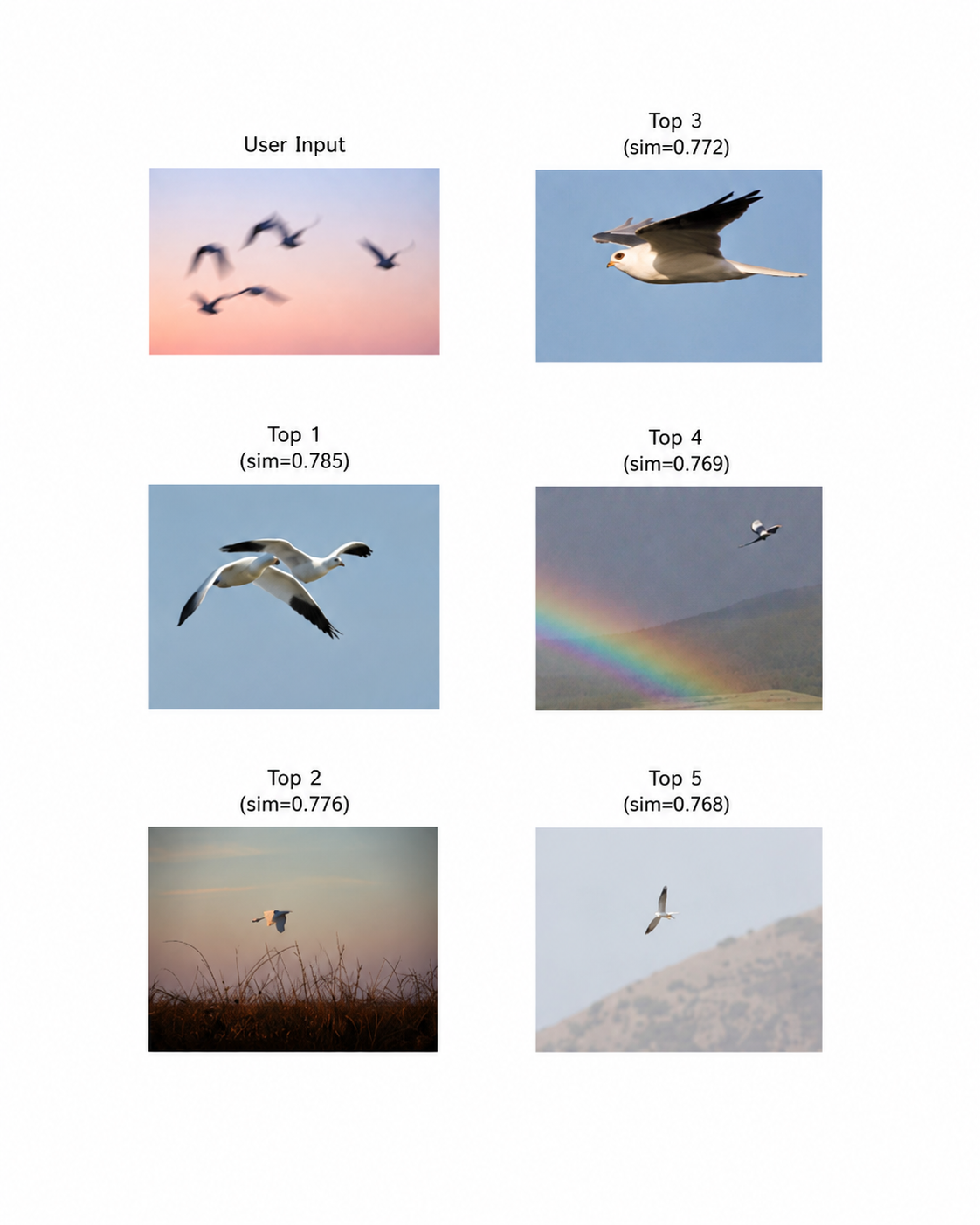}
\caption{\it Abstain/fallback qualitative case showing the input image and top-5 retrieved evidence for the birds-in-flight example.}
\label{fig:birds_evidence}
\end{figure}

\begin{figure}
\centering
\includegraphics[width=0.80\columnwidth,trim=0.03in 0.03in 0.03in 0.03in,clip]{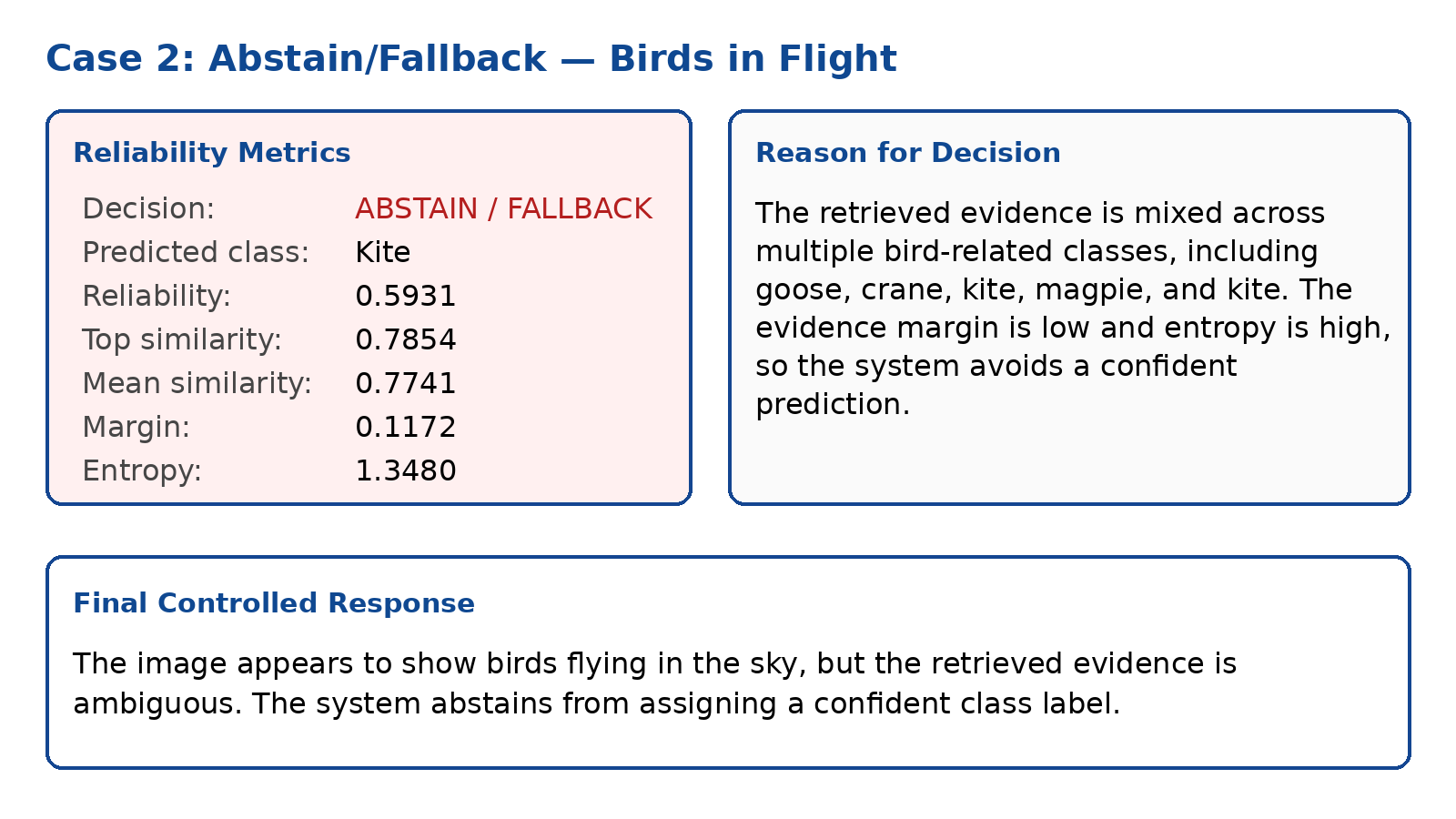}
\caption{\it Reliability summary and final controlled response for the birds-in-flight fallback example.}
\label{fig:birds_summary}
\end{figure}

\\[-12pt]
\noindent\textit{Case 2: Abstain/fallback--birds in flight.}
The second case, shown in Figure~\ref{fig:birds_evidence} and Figure~\ref{fig:birds_summary}, contains a group of blurred birds flying in the sky. The retrieval system predicts the class as kite. However, the top retrieved evidence is distributed across multiple bird-related classes, including goose, crane, kite, and magpie. The reliability score is 0.5931, the top similarity is 0.7854, the mean similarity is 0.7741, the evidence margin is 0.1172, and the entropy is 1.3480. The low margin and high entropy indicate class conflict among the retrieved examples, so the system assigns the case to ABSTAIN/FALLBACK.

\par\noindent

\par\noindent
\begin{figure}
\centering
\includegraphics[width=0.63\columnwidth,trim=0.03in 0.03in 0.03in 0.03in,clip]{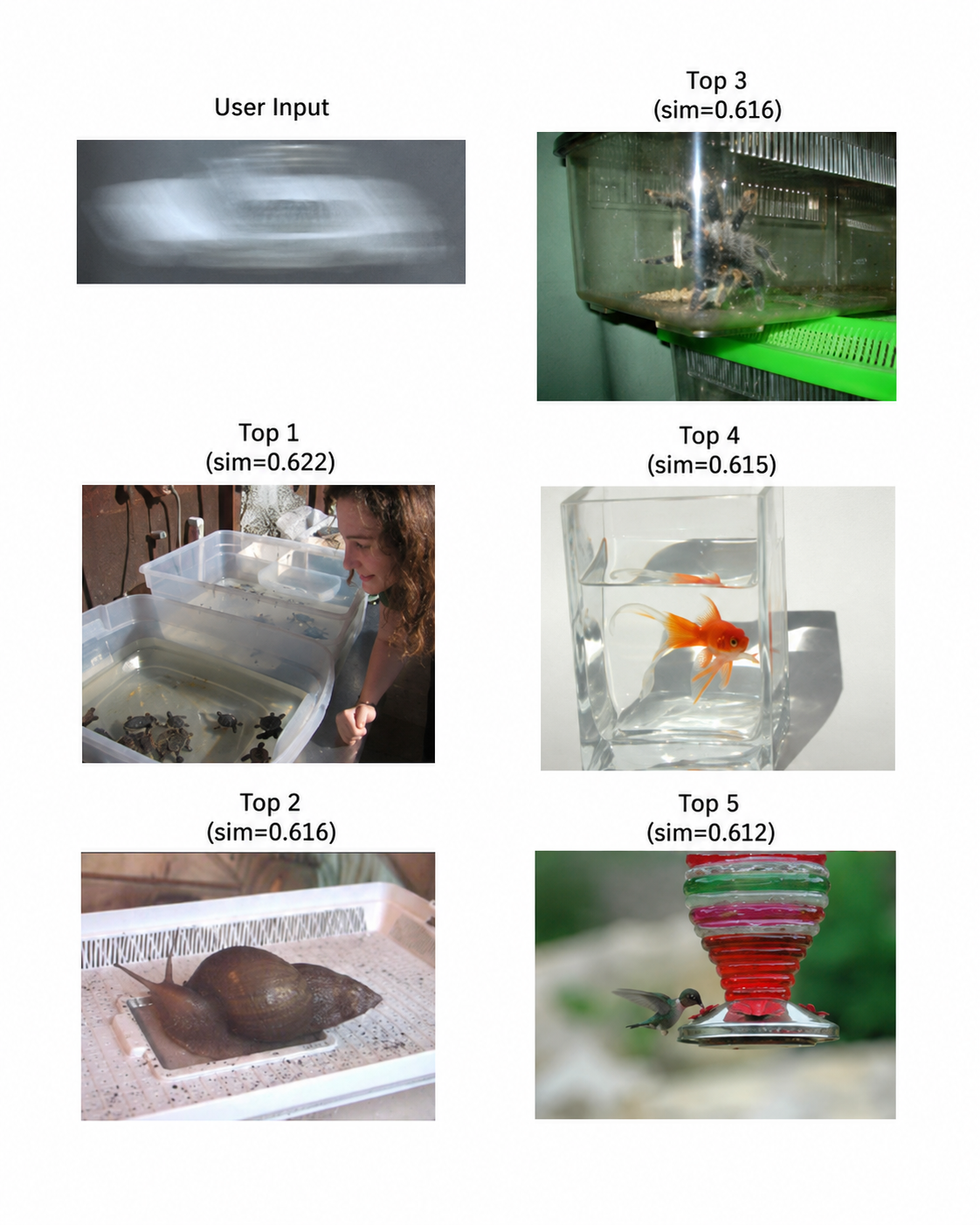}
\caption{\it Abstain/fallback qualitative case showing the input image and top-5 retrieved evidence for the blurred car example.}
\label{fig:blurred_car_evidence}
\end{figure}

\begin{figure}
\centering
\includegraphics[width=0.80\columnwidth,trim=0.03in 0.03in 0.03in 0.03in,clip]{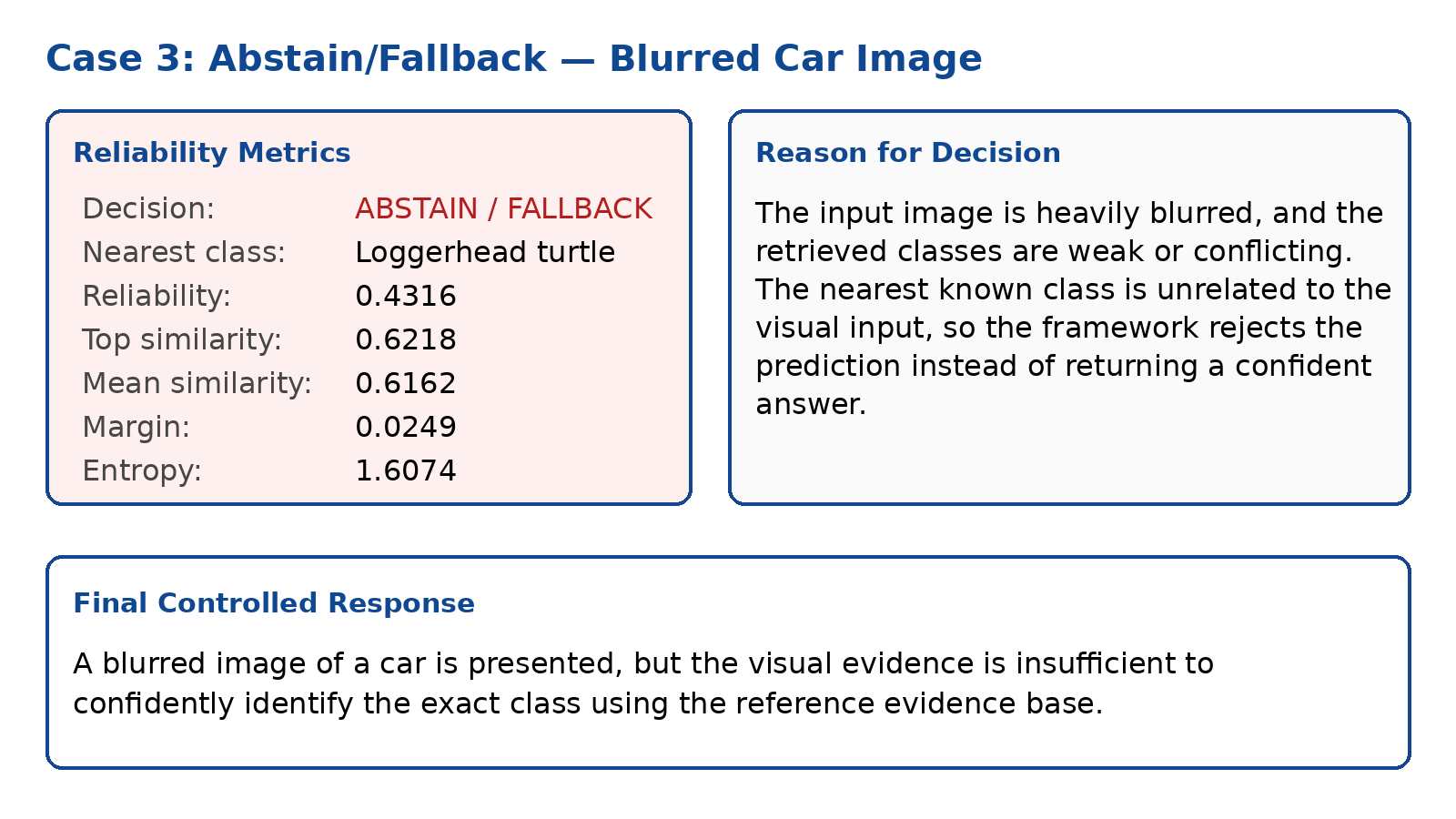}
\caption{\it Reliability summary and final controlled response for the blurred car fallback example.}
\label{fig:blurred_car_summary}
\end{figure}

\\[-12pt]
\noindent\textit{Case 3: Abstain/fallback--blurred car.}
The third case, shown in Figure~\ref{fig:blurred_car_evidence} and Figure~\ref{fig:blurred_car_summary}, contains a heavily blurred car image. The nearest retrieved class is loggerhead turtle, which is not semantically appropriate for the uploaded image. The retrieved evidence includes loggerhead turtle, snail, tarantula, goldfish, and hummingbird. The reliability score is 0.4316, the top similarity is 0.6218, the mean similarity is 0.6162, the evidence margin is 0.0249, and the entropy is 1.6074. These values indicate weak and conflicting evidence, so the system abstains instead of forcing an incorrect label.

\par\noindent

\par\noindent
\begin{figure}
\centering
\includegraphics[width=0.63\columnwidth,trim=0.03in 0.03in 0.03in 0.03in,clip]{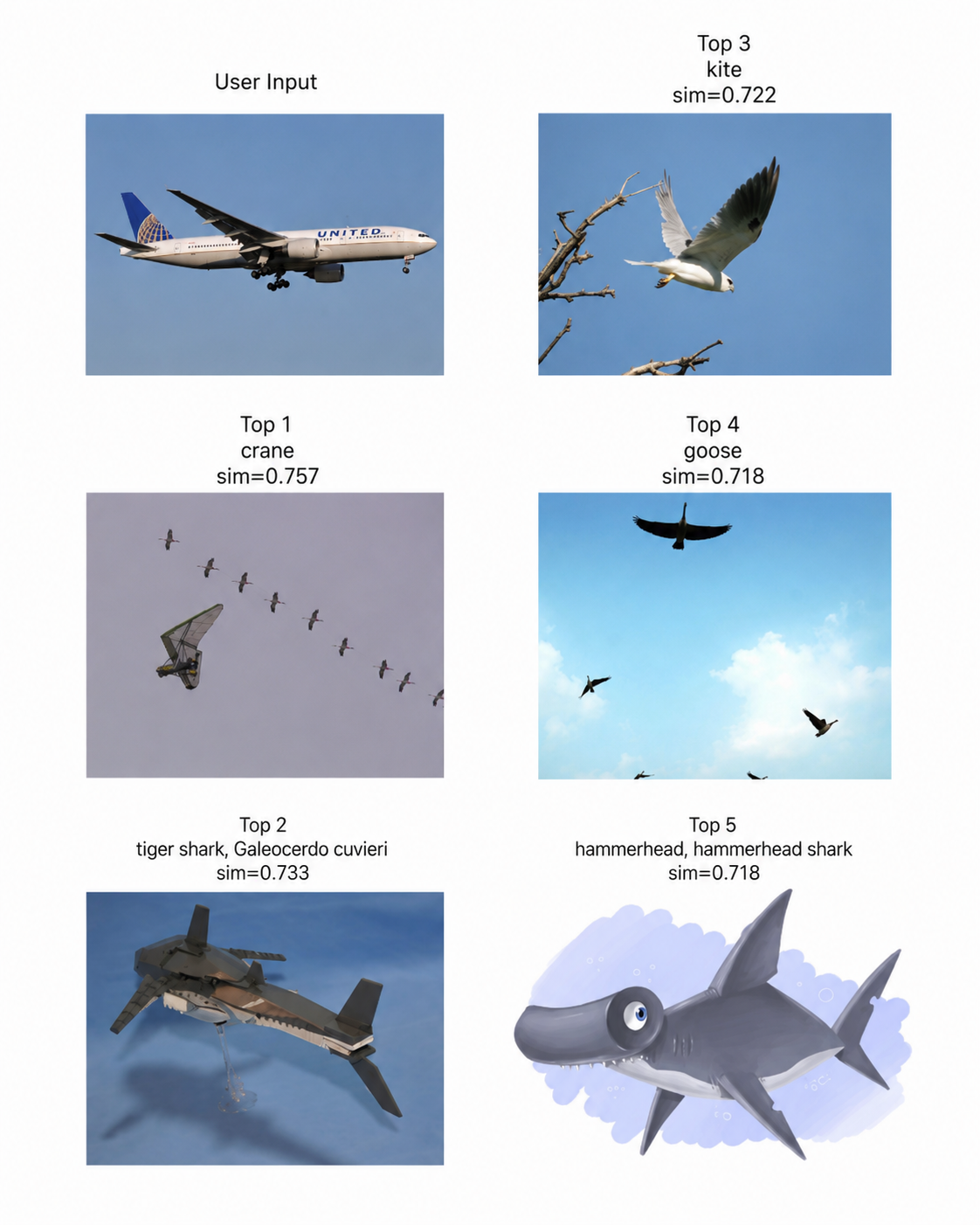}
\caption{\it Abstain/fallback qualitative case showing the input image and top-5 retrieved evidence for the airplane example.}
\label{fig:airplane_evidence}
\end{figure}

\begin{figure}
\centering
\includegraphics[width=0.80\columnwidth,trim=0.03in 0.03in 0.03in 0.03in,clip]{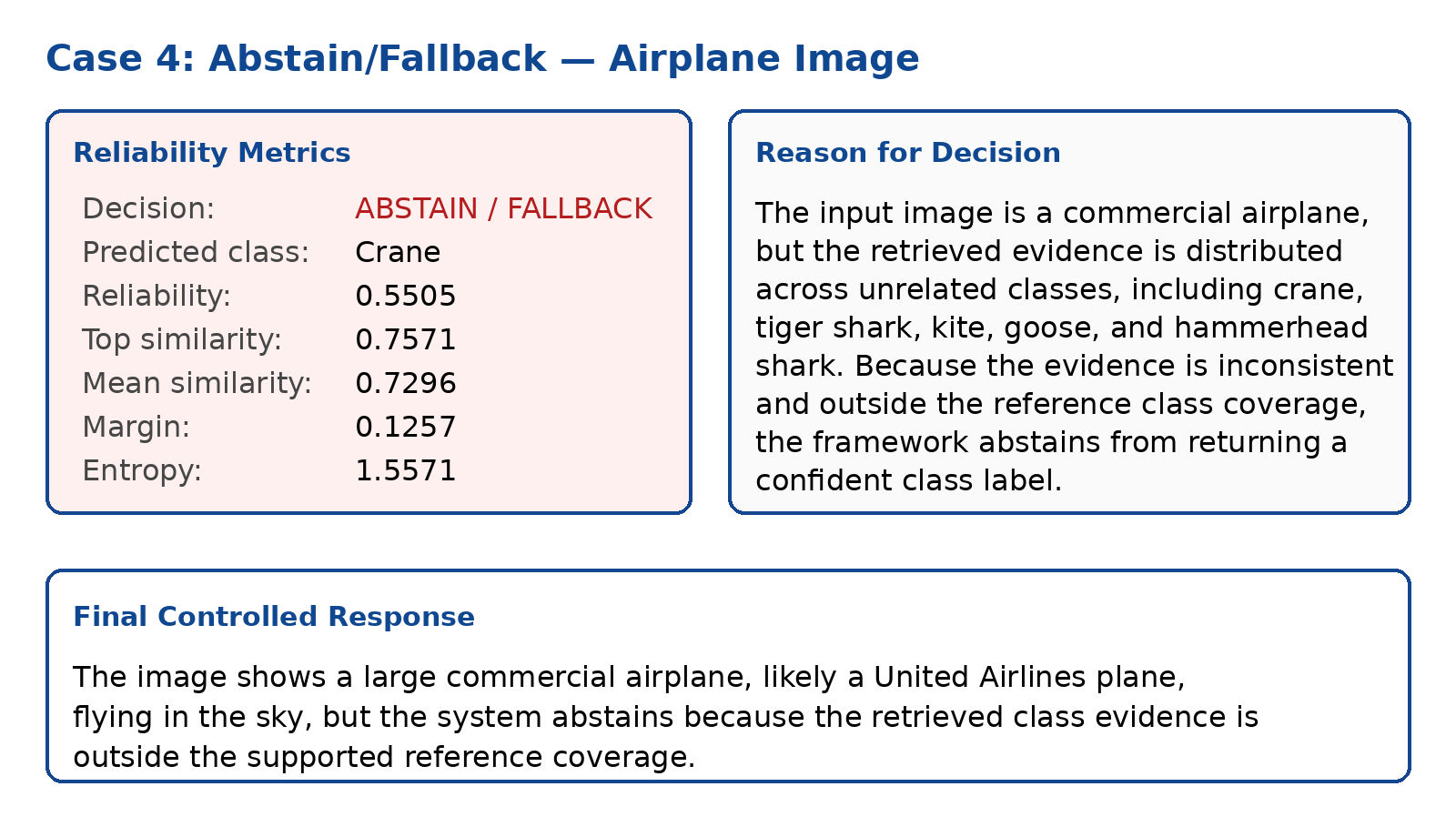}
\caption{\it Reliability summary and final controlled response for the airplane fallback example.}
\label{fig:airplane_summary}
\end{figure}

\\[-12pt]
\noindent\textit{Case 4: Abstain/fallback--airplane.}
The fourth case, shown in Figure~\ref{fig:airplane_evidence} and Figure~\ref{fig:airplane_summary}, contains a commercial airplane. The retrieval system predicts the class as crane because airplane is not supported by the ImageNet-100 reference subset used in this experiment. The top retrieved classes include crane, tiger shark, kite, goose, and hammerhead shark. The reliability score is 0.5505, the top similarity is 0.7571, the mean similarity is 0.7296, the evidence margin is 0.1257, and the entropy is 1.5571. Although the visual-language response can describe the image as an airplane, the retrieval evidence does not support a reliable ImageNet-100 class prediction, so the reliability gate assigns the case to ABSTAIN/FALLBACK.

\par\noindent

\par\noindent
\\[-12pt]
Overall, these qualitative examples demonstrate that the proposed framework behaves in a reliability-aware manner across different input conditions. The shark example shows acceptance under strong and consistent evidence. The birds example shows abstention under class ambiguity among visually related categories. The blurred car example shows fallback under low-quality and weak-evidence conditions. The airplane example shows protection against out-of-coverage inputs. Together, these cases show that the framework reduces overconfident visual hallucinations by preventing unsupported predictions from being accepted as final answers.
\subsection{Result Discussion}
The experimental results show that retrieval evidence and reliability-aware gating improve the trustworthiness of visual inference. The baseline retrieval model achieves an overall accuracy of 85.84\%, but it always returns an answer, including for weak-evidence, ambiguous, and out-of-coverage cases. This always-answering behavior is risky in user-facing multimodal systems because incorrect predictions may still be presented as confident outputs. The proposed framework addresses this issue by separating prediction generation from prediction acceptance.
\\
\\
The proposed reliability-aware framework improves accepted accuracy from 85.84\% to 88.88\%. This improvement occurs because the gate rejects a portion of samples where the retrieved evidence is weak, conflicting, or below the learned similarity thresholds. The system accepts 89.04\% of validation samples and abstains on 10.96\%. This result shows a reliability-coverage trade-off: the system gives up a small portion of coverage in order to increase the correctness of the predictions it actually returns.
\\
\\
The hallucination-like accepted wrong-answer rate decreases from 14.16\% to 11.12\%. This corresponds to an absolute reduction of 3.04 percentage points and a relative reduction of 21.48\%. This result is important because the main objective of the work is not simply to maximize raw classification accuracy, but to reduce accepted wrong predictions that may appear reliable to the user. In this work, such accepted wrong predictions are treated as hallucination-like visual errors because the system would otherwise provide an unsupported visual answer.
\\
\\
The calibration results further support the effectiveness of the reliability gate. ECE decreases from 7.40\% for the baseline retrieval system to 5.87\% for the proposed framework. The number of confident wrong cases also decreases from 263 to 226. These improvements indicate that the proposed reliability score provides a better estimate of when the system should trust its own prediction. In practical terms, the model becomes less likely to provide a highly confident answer when the evidence does not justify that confidence.
\\
\\
The evidence-quality analysis explains why the gate works. Accepted samples have higher average reliability, higher top similarity, higher mean similarity, larger evidence margin, and lower entropy. In contrast, abstained/fallback samples have lower accuracy, weaker similarity, smaller margins, and higher entropy. This shows that the gate is not rejecting samples randomly. Instead, it is rejecting cases that are empirically more difficult and less reliable. The accepted group achieves 88.88\% accuracy, while the abstained/fallback group has only 61.13\% accuracy. This large difference confirms that the reliability signals separate more trustworthy predictions from less trustworthy ones.
\\
\\
Overall, the proposed retrieval-augmented reliability-aware inference framework provides a practical post-hoc reliability layer for multimodal visual systems. It does not require retraining the underlying visual encoder or large multimodal model. Instead, it uses external evidence retrieval, class-support estimation, uncertainty measurement, and selective decision gating to control when a prediction should be accepted, softened, or rejected. This makes the framework suitable for trustworthy multimodal applications where reliable response behavior is more important than forcing a prediction for every input.

%%\endgroup

%% file: limitation.tex
\section{Limitation}
Although the proposed retrieval-augmented reliability-aware inference framework improves accepted accuracy and reduces overconfident visual errors, several limitations remain. First, the current implementation is evaluated on ImageNet-100, which provides a controlled visual classification setting but does not fully represent the open-world complexity of real multimodal systems. User-uploaded images may contain objects, scenes, or concepts that are not included in the reference evidence database. In such cases, the proposed framework can abstain or fall back, but it cannot assign a correct class if the target class is absent from the reference set.
\\
\\
Second, the framework depends on the quality and coverage of the external evidence database. If the reference database does not contain visually representative examples, the retrieved evidence may be weak or misleading. This limitation is visible in out-of-coverage examples where the system retrieves semantically unrelated classes. Therefore, the reliability of the framework can be improved by expanding the evidence database with broader and more diverse visual categories.
\\
\\
Third, the current reliability score is based on hand-designed evidence signals, including similarity strength, class-support agreement, margin, and entropy. Although these signals are interpretable and effective in the current experiment, the weighting strategy may not be optimal for every dataset or visual encoder. Future work can improve this by learning the reliability function from validation data using supervised calibration models or uncertainty-aware classifiers.
\\
\\
Fourth, the current framework uses ResNet-50 embeddings for visual evidence retrieval. While ResNet-50 provides efficient and stable visual representations, more powerful vision-language encoders such as CLIP, SigLIP, or multimodal embedding models may provide stronger semantic alignment between visual content and textual class descriptions. Replacing or combining the current visual encoder with vision-language embeddings may improve retrieval quality, especially for fine-grained or open-vocabulary visual categories.
\\
\\
Finally, the current response-generation layer is evaluated mainly as a controlled demonstration. The reliability gate controls whether a prediction should be accepted, softened, or rejected, but the final natural-language response can still be improved. In particular, future systems should ensure that the language model always follows the evidence-gated decision and does not overstate unsupported predictions. This is especially important for real-world multimodal applications where users may rely on the generated explanation as much as the predicted class label.

%% file: discussion_conclusion.tex
\section{Discussion and Conclusion}
This work presented a retrieval-augmented reliability-aware inference framework for reducing visual hallucinations in multimodal systems. The main idea of the proposed approach is to separate prediction generation from prediction acceptance. Instead of allowing the system to return a confident answer for every input image, the framework first checks whether the predicted class is supported by external visual evidence retrieved from a reference database.
\\
\\
The proposed method uses pretrained visual embeddings, FAISS-based nearest-neighbor retrieval, class-support aggregation, similarity strength, evidence margin, entropy-based uncertainty, and an aggregate reliability score to estimate instance-level trustworthiness. Based on these signals, the decision gate determines whether a prediction should be accepted, answered with caution, or rejected through abstention/fallback. This makes the system more reliable because the final response is controlled by evidence quality rather than by the predicted label alone.
\\
\\
The experimental results show that the proposed framework improves the reliability of accepted predictions and reduces hallucination-like accepted wrong answers. More importantly, the evidence-quality analysis demonstrates that accepted samples are supported by stronger similarity, larger margins, lower entropy, and more consistent retrieved evidence, while abstained samples are generally weaker and more ambiguous. This confirms that the decision gate is not rejecting inputs randomly, but is identifying cases where the visual evidence is insufficient for a confident answer.
\\
\\
The qualitative examples further highlight the practical value of the framework. When retrieved evidence is strong and class-consistent, as in the great white shark example, the system accepts the prediction. When the evidence is ambiguous, degraded, or outside the reference class coverage, as shown in the birds, blurred car, and airplane examples, the framework avoids forcing an unsupported answer. These cases demonstrate how retrieval evidence and reliability gating can prevent visually unsupported predictions from becoming confident user-facing responses.
\\
\\
Overall, this work shows that retrieval-augmented reliability estimation can serve as a practical post-hoc trust layer for multimodal visual inference. The framework does not require retraining large multimodal models, making it suitable for integration with existing visual encoders and multimodal response-generation systems. By combining external evidence retrieval, uncertainty estimation, and selective decision control, the proposed approach provides a step toward safer and more trustworthy multimodal AI systems.
\\
\\
Future work will extend this framework to larger and more diverse open-vocabulary datasets, stronger vision-language encoders, and learned reliability calibration methods. Additional improvements can also explore richer multimodal evidence sources, such as image-text retrieval, visual question answering feedback, and language-model-based explanation verification, to further improve the reliability of real-world multimodal reasoning systems.